\documentclass[sigconf,natbib=false]{acmart}

\AtBeginDocument{%
  }

\setcopyright{acmcopyright}

\copyrightyear{2023} 
\acmYear{2023} 
\setcopyright{acmlicensed}\acmConference[MM '23]{Proceedings of the 31st ACM International Conference on Multimedia}{October 29-November 3, 2023}{Ottawa, ON, Canada}
\acmBooktitle{Proceedings of the 31st ACM International Conference on Multimedia (MM '23), October 29-November 3, 2023, Ottawa, ON, Canada}
\acmPrice{15.00}
\acmDOI{10.1145/3581783.3611803}
\acmISBN{979-8-4007-0108-5/23/10}
 
\usepackage{graphicx}
\usepackage{amsmath}
\graphicspath{{./figs/}}
\DeclareGraphicsExtensions{.pdf,.jpg,.png}

\usepackage{cases}
\usepackage{mdwmath}
\usepackage{mdwtab}
\usepackage{array}
\usepackage{url}
\usepackage[ruled,vlined]{algorithm2e}
\usepackage{threeparttable}
\usepackage{color}
\usepackage{pifont}
\usepackage{wasysym}
\usepackage{multirow}
\usepackage{makecell}
\usepackage{tabu}

\usepackage{flushend}
\usepackage{float}
\usepackage{subfig}

\newcommand{\myPara}[1]{\vspace{0.02in}\noindent\textbf{#1}}

\newcommand{\ie}{\textit{i}.\textit{e}., }
\newcommand{\eg}{\textit{e}.\textit{g}., }
\def\etal{{\em et al.}}
\newcommand{\etc}{\textit{e}\textit{t}\textit{c}}

\makeatletter

\newcommand{\Rmnum}[1]{\expandafter\@slowromancap\romannumeral #1@}
\makeatother
\RequirePackage[
  datamodel=acmdatamodel,
  style=acmnumeric,
  ]{biblatex}

\begin{document}

\title{All in One: Exploring Unified Vision-Language Tracking with Multi-Modal Alignment}

\author{Chunhui Zhang}
\authornote{The first two authors contributed equally. This work was done during the visiting at the Hong Kong University of Science and Technology (Guangzhou).}
\affiliation{%
  \institution{Cooperative Medianet Innovation Center, Shanghai Jiao Tong University\\ CloudWalk Technology Co., Ltd}
  %\\ Hong Kong University of Science and Technology (Guangzhou)
  %\city{Shanghai}
  %\country{China}
  \country{}
  }
%\email{chunhui.zhang@sjtu.edu.cn}

\author{Xin Sun}
\authornotemark[1]
\affiliation{%
  \institution{Cooperative Medianet Innovation Center, Shanghai Jiao Tong University\\ CloudWalk Technology Co., Ltd }
  %\city{Shanghai}
  %\country{China}
  \country{}
  }
%\email{huntersx@sjtu.edu.cn}

\author{ Yiqian Yang}
\affiliation{%
  \institution{Northwestern Polytechnical University\\Shenzhen Research Institute of Big Data}
  %\city{Xi'an}
  %\country{China}
  \country{}
  }
%\email{frank.stuart@mail.nwpu.edu.cn}

\author{Li Liu}
\authornote{Li Liu is the corresponding author (avrillliu@hkust-gz.edu.cn).}
\affiliation{%
  \institution{Hong Kong University of Science and Technology (Guangzhou)}
  %\city{Guangzhou}
  %\country{China}
  \country{}
  }
%\email{avrillliu@hkust-gz.edu.cn}

\author{Qiong Liu}
%\author{Xi Zhou}
\affiliation{%
  \institution{CloudWalk Technology Co., Ltd}
  %\city{Shanghai}
  %\country{China}
  \country{}
  }
%\email{liuqiong@cloudwalk.com}

\author{Xi Zhou}
\affiliation{%
  \institution{CloudWalk Technology Co., Ltd}
  %\city{Shanghai}
  %\country{China}
  \country{}
  }
%\email{zhouxi@cloudwalk.cn}

\author{Yanfeng Wang}
\affiliation{%
  \institution{Cooperative Medianet Innovation Center, Shanghai Jiao Tong University\\Shanghai AI Laboratory}
  %\city{Shanghai}
  %\country{China}
  \country{}
  }
%\email{wangyanfeng@sjtu.edu.cn}

\renewcommand{\shortauthors}{Chunhui Zhang et al.}

\begin{abstract}
  Current mainstream vision-language (VL) tracking framework consists of three parts, \textit{i}.\textit{e}., a visual feature extractor, a language feature extractor, and a fusion model. To pursue better performance, a natural modus operandi for VL tracking is employing customized and heavier unimodal encoders, and multi-modal fusion models. Albeit effective, existing VL trackers separate feature extraction and feature integration, resulting in extracted features that lack semantic guidance and have limited target-aware capability in complex scenarios, \textit{e}.\textit{g}., similar distractors and extreme illumination. In this work, inspired by the recent success of exploring foundation models with unified architecture for both natural language and computer vision tasks, we propose an All-in-One framework, which learns joint feature extraction and interaction by adopting a unified transformer backbone. Specifically, we mix raw vision and language signals to generate language-injected vision tokens, which we then concatenate before feeding into the unified backbone architecture. This approach achieves feature integration in a unified backbone, removing the need for carefully-designed fusion modules and resulting in a more effective and efficient VL tracking framework. To further improve the learning efficiency, we introduce a multi-modal alignment module based on cross-modal and intra-modal contrastive objectives, providing more reasonable representations for the unified All-in-One transformer backbone. Extensive experiments on five benchmarks, \textit{i}.\textit{e}., OTB99-L, TNL2K, LaSOT, LaSOT$_{\rm Ext}$ and WebUAV-3M, demonstrate the superiority of the proposed tracker against existing state-of-the-art (SOTA) methods on VL tracking. Codes will be available at \href{https://github.com/983632847/All-in-One}{\color{magenta}{here}}.
\end{abstract}





\maketitle

\begin{figure}[t]
  \centering
  \includegraphics[width=\linewidth]{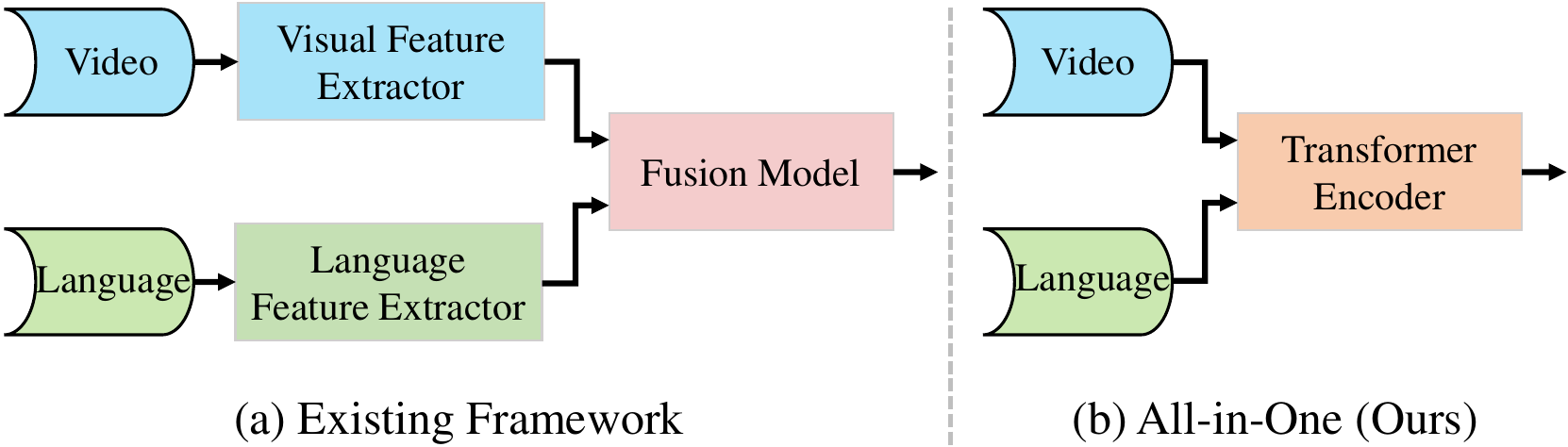}
  \vspace{-0.3cm}
  \caption{Existing VL tracking framework \emph{vs.} our All-in-One.  (a) Existing VL tracking methods obtain multiple modality features from separate extractors before fusion. The feature interaction relies on a carefully-designed fusion model. (b) We aim to build a foundation model, \ie All-in-One, for VL tracking, which achieves joint feature extraction and multi-modal interaction using a versatile transformer encoder.}
  \label{fig:Motivation}
\end{figure}

\section{Introduction}
Vision-language (VL) tracking~\cite{zhang2022webuav,wang2021towards,li2017tracking,feng2021siamese,guo2022divert}, one of the fundamental and challenging problems {at the intersection of computer vision and natural language understanding}, aims to locate the object in each frame of a video based on a given natural language prompt and an initial object box. It plays a crucial role in human-machine interaction, transportation surveillance, virtual reality, autonomous driving, delivery, \etc. Compared with traditional visual object tracking~\cite{huang2019got,wu2015otb} using a bounding box to describe the object of interest, VL tracking has the potential to achieve more robust tracking by leveraging the complementary superiority of multiple modalities.

In the past few years, two-stream VL trackers~\cite{wang2021towards,feng2021siamese,ma2021capsule,guo2022divert}, which extract visual features and language features separately and then perform feature interaction in a fusion model (as shown in Fig~\ref{fig:Motivation}(a)), have emerged as a domain framework and obtained significant progress. For instance, Feng \etal~\cite{feng2021siamese} proposed a Siamese natural language region proposal network for multi-stage feature extraction, and then applied an aggregation module to dynamically combine predictions from both visual and language modalities. Guo \etal~\cite{guo2022divert} suggested an asymmetrical modeling architecture to learn adaptive VL representations. Following the two-stream pipeline, the latest transformer-based VL tracker JointNLT~\cite{zhou2023joint} formulates grounding and tracking as a unified task of establishing relation between visual-language references and test images via a multi-source relation modeling architecture.

Despite the convincing designs of existing two-stream VL trackers, they still suffer from the fundamental challenge of learning target-aware capability in some complex and corner scenarios, \eg similar distractors, occlusion, and extreme illumination~\cite{zhang2022webuav,guo2022divert}. Firstly, the separation of feature extraction and integration prevents the model from performing early multi-modal feature interaction, resulting in limited object-background discriminative power~\cite{ye2022joint,chen2022backbone}. Although some works have attempted to design complicated~\cite{ma2021capsule} or multi-stage~\cite{feng2021siamese,guo2022divert} fusion models to enhance the associations between modalities, the lack of mutual interaction remains an insurmountable gap. More seriously, heavy fusion models increase the number of parameters, leading to significant computational inefficiency. Secondly, performing feature interaction directly ignores the huge distribution discrepancies between the vision and language modalities in the feature space~\cite{He2022MAE}, leading to significant learning inefficiency in VL representation learning.

To tackle the above issues, we propose a unified framework (as shown in Fig~\ref{fig:Motivation}(b)), namely All-in-One, for multi-modal VL tracking. The core idea is to establish bidirectional information flow between well-aligned visual and language signals as early as possible via a unified transformer backbone. Our All-in-One framework brings multiple new benefits for multi-modal VL tracking. (1) The unified architecture not only simplifies the model, but also leads to more efficient VL representation learning. (2) It has great potential to serve as a foundation model for VL tracking. With this framework, we develop a general VL tracking model, which generalizes well to complex, and user-defined language descriptions/prompts on various VL tracking datasets. (3) Compared with the two-stream vision language foundation models such as CLIP~\cite{radford2021learning}, our All-in-One framework follows the simple and general one-stream architecture~\cite{ye2022joint,chen2022backbone,lin2022swintrack}.

Specifically, we introduce a versatile All-in-One transformer, as shown in Fig.~\ref{fig:framework}, to embed raw visual and language signals into joint VL representations, and the produced visual search region features can be directly used for object location without an additional fusion model. The visual inputs (\ie search region and template) and language input are first mapped by a patch embed and a text tokenizer, respectively, and then flattened into the same dimension embeddings. A modal mixup operation is used to inject language information into the visual embeddings (\ie template embeddings and search region embeddings), followed by a stack of transformer encoder layers enabling iteratively feature integrating between template and search region embeddings with language information guidance. Thus, both template and search region embeddings can be enhanced dynamically with strong target-aware capability. In addition, we introduce a multi-modal alignment (MMA) module to alleviate the huge distribution discrepancies between multiple modalities based on contrastive learning (CL)~\cite{oord2018representation}. The MMA module includes cross-modal alignment (CMA) and intra-modal alignment (IMA), forcing the visual and language signals from the same video to be close in the feature space, while making the distribution of multi-modal features more uniform and reasonable in the entire feature space, which can promote feature integration. %In conclusion, our main contributions can be summarized as follows:

Our main contributions can be summarized as three folds. First, we propose a simple, compact, and effective one-stream framework for VL tracking, namely All-in-One, which learns VL representations from raw visual and language signals end-to-end in a unified transformer backbone. Second, we develop a novel multi-modal alignment module incorporating cross-modal and intra-modal alignments to enable efficient multi-modal learning by aligning multiple signals in the feature space before learning. Third, we conduct extensive experiments and comparisons to demonstrate that our approach achieves higher accuracy against state-of-the-arts.

%\begin{itemize}
%\item We propose a simple, compact, and effective one-stream framework for VL tracking, namely All-in-One, which learns VL representations from raw visual and language signals end-to-end in a unified transformer backbone.

%\item We develop a novel multi-modal alignment module incorporating cross-modal and intra-modal alignments to enable efficient multi-modal learning by aligning multiple signals in the feature space before learning.

%\item Extensive experiments demonstrate that the proposed approach achieves higher accuracy against state-of-the-arts.
%\end{itemize}

\section{Related Work}
\subsection{Vision-Language Tracking}
In recent years, the two-stream  framework~\cite{wang2021towards,feng2021siamese,ma2021capsule,guo2022divert} has emerged as a dominant VL tracking paradigm (see Fig.~\ref{fig:Motivation}(a)). They first extract features using two independent unimodal feature extractors, and then model the relation of visual features and language features sequentially by a lightweight~\cite{guo2022divert} or relatively heavy~\cite{zhou2023joint} network. Early work~\cite{li2017tracking} contains a visual specification network and a lingual specification network, and further selectively focuses on parts of language prompts using a lingual specification attention network. Later, GTI~\cite{yang2020grounding} and~\cite{wang2021towards} decompose the VL tracking problem into three sub-tasks of visual tracking, grounding, and integration. VLT$_{\rm TT}$~\cite{guo2022divert} suggests learning VL representations through an asymmetrical modeling architecture. JointNLT~\cite{zhou2023joint} introduces a joint visual grounding and tracking framework by localizing the referred object based on the visual-language references. However, these works rely on separate visual and language encoders to extract multi-modal features, leading to limited information interaction. We note that several works~\cite{ye2022joint,chen2022backbone,lin2022swintrack} introduce a one-stream framework for visual object tracking. Different from them, we extend the one-stream framework to multi-modal VL tracking by training
jointly on videos and language prompts. As shown in Fig.~\ref{fig:Motivation}(b), for the first time we seamlessly integrate feature extraction and interaction into a unified backbone architecture for VL tracking. The proposed framework not only enables information flow from language to vision, but also allows bidirectional integration of information between visual and language features.

\begin{figure*}[t]
  \centering
 \includegraphics[width=0.9\linewidth]{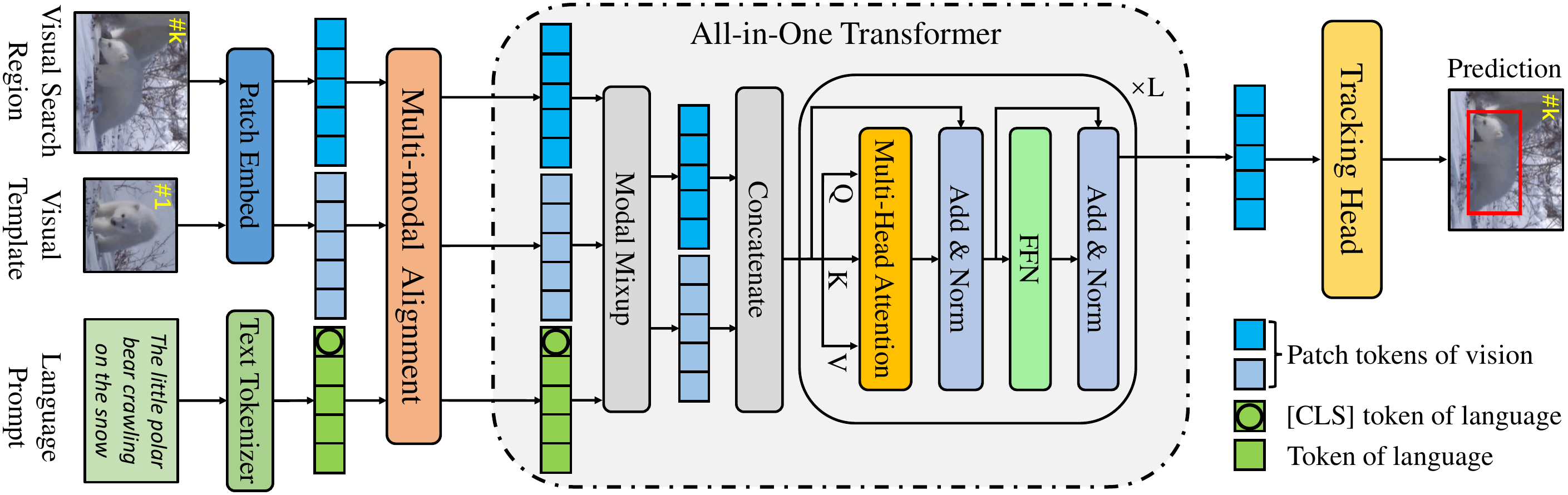}
  %\vspace{-0.3cm}
  \caption{Overview of the proposed All-in-One framework. The multi-modal alignment module is introduced before the All-in-One transformer backbone to align visual and language embeddings in the feature space. The All-in-One transformer backbone is applied to achieve joint feature extraction and interaction. The tracking head is used to predict object location.}
  \label{fig:framework}
\end{figure*}

\subsection{Transformer for Unified Architecture}
Thanks to the scalability to very large-scale model and capability to handle sequential and non-sequential data, transformer has become a prevailing architecture in both natural language~\cite{vaswani2017attention,devlin2018bert} and computer vision~\cite{dosovitskiy2020image,carion2020end} communities. Following ViT~\cite{dosovitskiy2020image}, a series of variants of ViT have been developed to improve the performance on vision tasks, including reducing computational cost~\cite{touvron2021training,carion2020end}, and architecture design~\cite{wang2021pyramid,chen2021crossvit}. Additionally, transformer has been extensively used in various multi-modal tasks~\cite{wang2019reinforced,salvador2021revamping,zhang2022vldeformer}.

In consideration of the capacity of the unified transformer model to handle unimodal input or multi-modal input with a shared encoder, a few pioneering works have tried to explore unified multi-modal encoders~\cite{wang2022all,kim2021vilt,akbari2021vatt}. For instance, ViLT~\cite{kim2021vilt} proposed a vision-language transformer without using regional features or deep convolutional visual embeddings. VATT~\cite{akbari2021vatt} developed a video-audio-text transformer using multi-modal self-supervised pre-training to improve the performance of video action recognition and audio event classification. In this paper, we follow the trend of unified architecture for multi-modal data. To the best of our knowledge, the proposed All-in-One transformer is the first unified backbone architecture for multi-modal VL tracking.

\subsection{Multi-Modal Learning}
Recently, a potential multi-modal learning paradigm is to adopt transformer to process and relate information from multiple
modalities~\cite{baltruvsaitis2018multimodal,li2019visualbert,Lee_2023_CVPR,CoMAE2022Yang, Bachmann2022MultiMAE}. CLIP~\cite{radford2021learning} applied language prompts as supervisory signals to learn better visual representations. VisualBERT~\cite{li2019visualbert}, VilBERT~\cite{lu2019vilbert}, and Unicoder-VL~\cite{li2020unicoder} combined visual and textual features into transformers to capture cross-modal relationships. However, previous works mainly focus on how to learn multi-modal representations by exploiting the complementary advantages of multiple modalities, or to fuse multi-modal features for prediction. Multi-modal alignment, discovering relationships and correspondences between fine-grained elements (\eg objects and words) of instances (\eg images and languages) from two or more modalities, has been rarely explored. In this work, we propose the MMA module with CMA and IMA based on self-supervised CL~\cite{oord2018representation} to explore efficient multi-modal learning for VL tracking.

\section{Proposed Method}
This section presents the All-in-One, a simple yet effective framework for the VL tracking task. Our All-in-One framework consists of an All-in-One transformer backbone, a multi-modal alignment module, and a tracking head, as shown in Fig.~\ref{fig:framework}. The All-in-One transformer backbone is used to achieve feature extraction and information interaction between visual inputs (\ie visual search region and visual template) and language input simultaneously in a unified architecture. Before that, visual embeddings and language embeddings are aligned via a multi-modal alignment module, providing more reasonable feature embeddings in the feature space. The output features of the visual search region are sent to the tracking head to predict the location of the target.

\subsection{Problem Formulation} 
Before detailing the architecture of our All-in-One framework, we briefly review the transformer tracking~\cite{chen2022backbone,cui2022mixformer,ye2022joint,chen2021transformer,lin2022swintrack}, which achieves remarkable tracking performance. Given a video with a pair of visual template and visual search region $\mathcal{X}_{xz}$, an initial target box $\mathcal{B}_{0}$, the transformer tracking can be formulated as $F_{trans}:\{\mathcal{X}_{xz},\mathcal{B}_{0}\}\rightarrow \mathcal{B}$, where $F_{trans}$ is the transformer tracker, $\mathcal{B}$ is the predicted box of the target in all subsequent search frames. In general, the transformer tracker $F_{trans}$ can be decomposed into $\Phi \circ f$, where $f:\{\mathcal{X}_{xz},\mathcal{B}_{0}\}\rightarrow \mathcal{H} $ denotes the backbone (\eg ViT~\cite{dosovitskiy2020image}) for feature extraction and interaction function, $\mathcal{H}$ represents the output features of visual search region, and the tracking head $\Phi:\mathcal{H}\rightarrow \mathcal{B}$ is adopted to predict the target box.

Specifically, a pair of images, namely visual search region $\mathbf{x} \in \mathbb{R}^{3 \times H_{x} \times W_{x}}$ and visual template $\mathbf{z} \in \mathbb{R}^{3 \times H_{z} \times W_{z}}$ are divided into $N_{x}$ and $N_{z}$ non-overlapping image patches of resolution $P \times P$, where $N_{x}=H_{x}W_{x}/P^{2}$ and $N_{z}=H_{z}W_{z}/P^{2}$ are the number of patches of search region and template, respectively. Then, a linear projection is applied to these image patches to generate 1D tokens $\mathcal{H}_{x} \in \mathbb{R}^{N_{x} \times D}$ and $\mathcal{H}_{z} \in \mathbb{R}^{N_{z} \times D}$, where $D$ is the token dimension. Two learnable positional embeddings are added to $\mathcal{H}_{x}$ and $\mathcal{H}_{z}$ to retain the position information. After that, these tokens are concatenated as a sequence    $\mathcal{H}_{xz}^{0} = [\mathcal{H}_{x}; \mathcal{H}_{z}]$ and fed to a $L$-layer transformer encoder. Here, we represent $\mathcal{H}_{xz}^{l-1}$ as inputs to the $l$-th encoder layer $E^{l}$. Formally, the forward operation of the $l$-th encoder layer can be written as:
%\vspace{-0.3cm}
\begin{equation}
  \mathcal{H}^{l}_{xz}=E^{l}(\mathcal{H}^{l-1}_{xz}),~~~~~~~~~~~~~~~~~~~~~~~~~~~~~~~~   l = 1,2,3,..., L
  \label{eq:extractor}
\end{equation}
\vspace{-0.3cm}
\begin{equation}
\mathcal{B}=\Phi(\mathcal{H}^{L}_{xz}),
  \label{eq:tracking_head}
\end{equation}
where each transformer encoder layer contains a multi-head self-attention (MHSA), and a feed-forward network (FFN). Each sub-layer is constructed as a residual connection, where layer normalization (LN) is followed by the residual connection. The visual search region tokens $\mathcal{H}^{L}_{x}$ of the last transformer encoder layer is taken as the input of tracking head $\Phi$ for target box prediction. 

For VL tracking~\cite{zhang2022webuav,fan2019lasot,wang2021towards}, it introduces an extra language prompt $\mathcal{T}$ for each video to express the attribute, behavior, position (relative location), and surroundings of the target. Accordingly, VL tracking can be formulated as $F_{VL}:\{\mathcal{X}_{xz},\mathcal{B}_{0},\mathcal{T}\}\rightarrow \mathcal{B}$, where $F_{VL}$ is the VL tracker. Similarly, the VL tracker $F_{VL}$ can also be decomposed into $\Phi \circ f^{*}$, where $\Phi$ is the tracking head, and $f^{*}$ represents the proposed unified backbone architecture in this work.

\subsection{Unified Vision-Language Tracking}

Fig.~\ref{fig:framework} gives an overview of our All-in-One framework for VL tracking. To optimize the VL tracker $F_{VL}$, a pair of visual template and visual search region $\mathcal{X}_{xz}=\{\mathcal{X}_{x}, \mathcal{X}_{z}\}$, and an extra language prompt $\mathcal{T}$ are first fed to a patch embed (\ie a linear projection) and a text tokenizer~\cite{devlin2018bert}, respectively. They are mapped and flattened into $D$-dimension embeddings, where $D=768$. We denote them as vision tokens (\ie $\mathcal{H}^{0}_{x}$ and  $\mathcal{H}^{0}_{z}$), where $\mathcal{H}^{0}_{x}\in\mathbb{R}^{N_{x}\times D}$ and  $\mathcal{H}^{0}_{z}\in\mathbb{R}^{N_{z}\times D}$ are visual search region tokens and visual template tokens, and language tokens $\mathcal{H}^{0}_{t}\in\mathbb{R}^{N_{t}\times D}$, where $N_{t}$ is the number of language tokens. Following~\cite{devlin2018bert}, a special classification token ([CLS]) is attached at the beginning of the language tokens. Then, $\mathcal{H}^{0}_{x}$,  $\mathcal{H}^{0}_{z}$, and $\mathcal{H}^{0}_{t}$ are aligned with the multi-modal alignment module (see section~\ref{sec:mma}) in the embedding space. It is worth noting that the well-aligned vision embeddings and language embeddings can facilitate multi-modal representation learning and interaction~\cite{baltruvsaitis2018multimodal}. Here, we still refer to the aligned vision embeddings and language embeddings as $\mathcal{H}^{0}_{x}$,  $\mathcal{H}^{0}_{z}$, and $\mathcal{H}^{0}_{t}$, respectively. Afterward, we perform a modal mixup operation~\cite{guo2022divert} between the aligned vision embeddings and language embeddings as follows:
\begin{equation}
\mathbf{F}^{0}_{x}=\mathcal{H}^{0}_{x}\odot Linear(\mathcal{H}^{0}_{t})+\mathcal{H}^{0}_{x},
  \label{eq:mixupx}
\end{equation}
\begin{equation}
\mathbf{F}^{0}_{z}=\mathcal{H}^{0}_{z}\odot Linear(\mathcal{H}^{0}_{t})+\mathcal{H}^{0}_{z},
  \label{eq:mixupz}
\end{equation}
where $\odot$ represents the Hadamard product, $Linear(\cdot)$ is a linear projection layer. In this way, the language information is injected into vision embeddings via the modal mixup operation. Moreover, Eqs.~(\ref{eq:mixupx})-(\ref{eq:mixupz}) also construct a bidirectional information ﬂow between vision and language modalities that allows mutual guidance for multi-modal feature extraction and interaction. By establishing a bidirectional information flow between well-aligned visual and language signals as early as possible via a unified transformer backbone, we can avoid the loss of discriminative information and thus make the extracted features highly target-aware~\cite{ye2022joint}.

Formally, the operations of the $l$-th encoder of our All-in-One transformer backbone can be expressed as:
\begin{equation}
\mathbf{Q}=\mathbf{K}=\mathbf{V}=[\mathbf{F}^{l}_{x}; \mathbf{F}^{l}_{z}],
  \label{eq:qkv}
\end{equation}
\begin{equation}
[\mathbf{F}^{\prime l}_{x}; \mathbf{F}^{\prime l}_{z}]={\rm LN}([\mathbf{F}^{l}_{x}; \mathbf{F}^{l}_{z}]+{\rm MHSA}(\mathbf{Q},\mathbf{K},\mathbf{V})),
  \label{eq:mhsa}
\end{equation}
\begin{equation}
[\mathbf{F}^{l+1}_{x}; \mathbf{F}^{l+1}_{z}]={\rm LN}([\mathbf{F}^{\prime l}_{x}; \mathbf{F}^{\prime l}_{z}]+{\rm FFN}([\mathbf{F}^{\prime l}_{x};\mathbf{F}^{\prime l}_{z}])),
  \label{eq:ffn}
\end{equation}
where $\mathbf{Q}$, $\mathbf{K}$, and  $\mathbf{V}$ represent the query, key, and value embeddings, $[;]$ denotes the concatenation operation, $\mathbf{F}^{l}_{x}$ and $\mathbf{F}^{l}_{z}$ are the input embeddings of the $l$-th transformer encoder. Therefore, the language information-injected vision embeddings are jointly processed by the transformer encoder, enabling seamless multi-modal feature extraction and integration. Finally, the visual search region embeddings of the last layer of the transformer encoder are reshaped into a 2D feature map. The feature map is fed into the tracking head to predict the location of the target.

\begin{figure}[t]
  \centering
  \includegraphics[width=\linewidth]{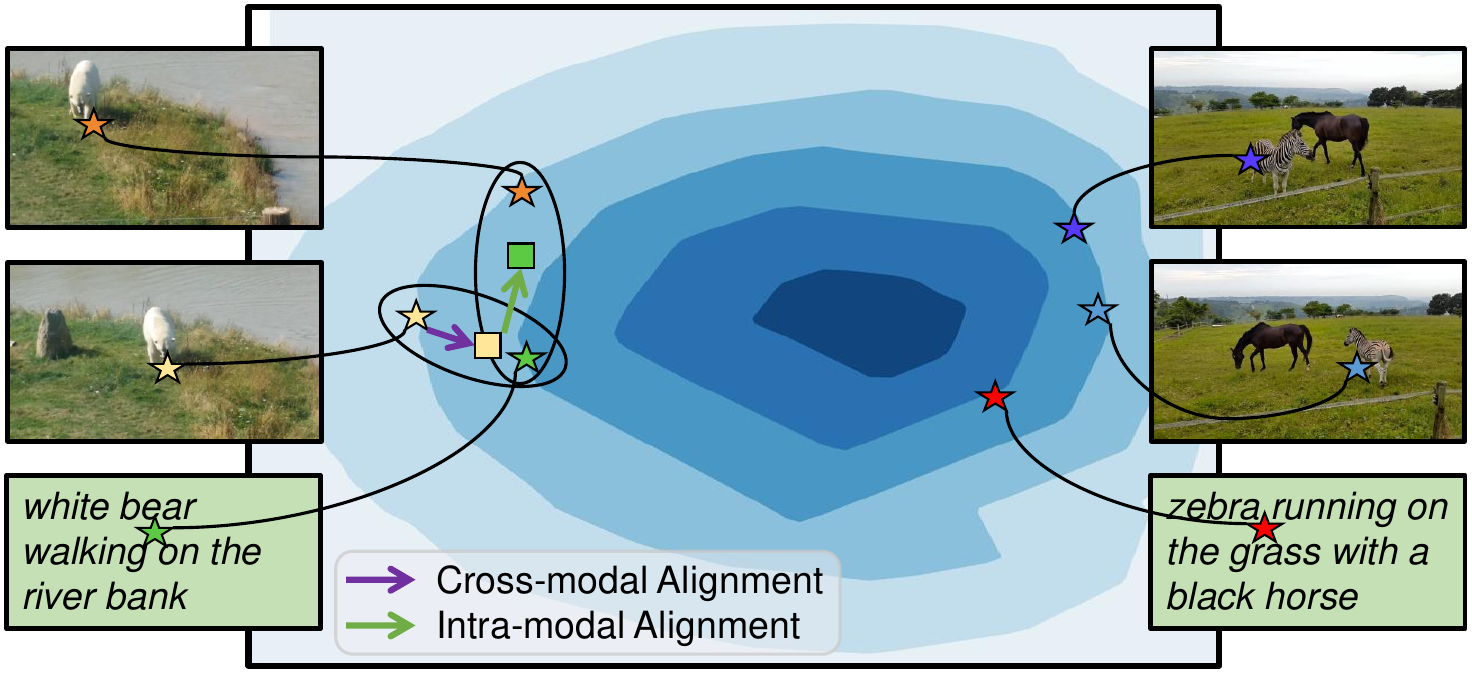}
  \vspace{-0.3cm}
  \caption{Illustration of MMA module, which contains CMA and IMA. For CMA only, the second vision embedding (yellow star) is pulled towards its matched language embedding (green star). By incorporating IMA, it can learn more reasonable embedding (yellow square to green square).}
  \label{fig:MMA}
\end{figure}

To model the interaction between language and vision features, recent VL trackers~\cite{zhou2023joint,guo2022divert} adopted a customized fusion model to directly serialize vision and language embeddings into sequences to learn a joint multi-modal embedding. Although, our All-in-One transformer backbone using the pretrained ViT ~\cite{dosovitskiy2020image} has the ability to model long-range dependencies of sequential data, alleviating the negative effects of modal differences for multi-modal learning, the vision embeddings and language embeddings lying in different feature spaces is still challenging for the transformer encoder to learn their interactions~\cite{li2021align,yang2022vision}. To tackle this limitation, we further propose a self-supervised MMA module, which is used before feature extraction and integration. The MMA module includes CMA and IMA, which can efficiently learn more reasonable feature distributions as shown in Fig.~\ref{fig:MMA}.

\subsection{Multi-Modal Alignment Module}
\label{sec:mma}

\myPara{Cross-modal Alignment.}
Since the vision and language embeddings from the same video are distributed in different feature spaces, a natural thought is to enforce them close in the feature space to reduce the difficulty for multi-modal interaction. Absorb this in mind, we introduce the CMA to pull the matched vision and language embeddings closer in feature space, while pushing away mismatched pairs. Actually, the goal of CMA is to maximize the mutual information (MI)~\cite{oord2018representation} between vision and language that are matched, which contain the same semantics. Fig.~\ref{fig:MMA} presents an example, the high-level language embedding (\ie green star) and sparse vision embedding (\ie yellow star) from the same video are pulled closer in the feature space. Specifically, visual search region tokens $\mathcal{H}^{0}_{x}$, visual template tokens $\mathcal{H}^{0}_{z}$ and language tokens $\mathcal{H}^{0}_{t}$ are projected into the same dimension through three linear projections, which we denote as $\mathbf{f}_{x}\in \mathbb{R}^{C}$,  $\mathbf{f}_{z}\in \mathbb{R}^{C}$, and $\mathbf{f}_{t}\in \mathbb{R}^{C}$, respectively, where $C=256$. To maximize the MI of vision and language tokens, we optimize the InfoNCE loss~\cite{oord2018representation} between vision and language, denoting the lower bound of their MI. Formally, InfoNCE losses of vision-to-language are defined as:
\begin{equation}
    \mathcal{L}_{x2t}(\mathbf{f}^{i}_{x},\mathbf{f}^{i}_{t},\widetilde{\mathbf{f}}_{t})=-\mathbb{E}_{(\mathbf{f}_{x},\mathbf{f}_{t})}\lbrack\log \frac{\exp(sim(\mathbf{f}^{i}_{x},\mathbf{f}^{i}_{t})/\tau)}{\sum_{j=1}^{N-1}\exp(sim(\mathbf{f}^{i}_{x},\widetilde{\mathbf{f}}^{j}_{t})/\tau)}\rbrack,
    \label{eq:x2t_loss}
\end{equation}
\vspace{-0.2cm}
\begin{equation}
    \mathcal{L}_{z2t}(\mathbf{f}^{i}_{z},\mathbf{f}^{i}_{t},\widetilde{\mathbf{f}}_{t})=-\mathbb{E}_{(\mathbf{f}_{z},\mathbf{f}_{t})}\lbrack\log \frac{\exp(sim(\mathbf{f}^{i}_{z},\mathbf{f}^{i}_{t})/\tau)}{\sum_{j=1}^{N-1}\exp(sim(\mathbf{f}^{i}_{z},\widetilde{\mathbf{f}}^{j}_{t})/\tau)}\rbrack,
    \label{eq:z2t_loss}
\end{equation}
where $\mathbf{f}^{i}_{x}$, $\mathbf{f}^{i}_{z}$, and $\mathbf{f}^{i}_{t}$ are two vision tokens and language tokens of the same video, respectively, $\widetilde{\mathbf{f}}_{t}=\{\widetilde{\mathbf{f}}^{1}_{t},...,\widetilde{\mathbf{f}}^{N-1}_{t}\}$ is a set of negative language examples for $\mathbf{f}^{i}_{x}$ or $\mathbf{f}^{i}_{z}$, $N$ is the batch size, $sim(\mathbf{f}^{i}_{x}, \mathbf{f}^{i}_{t})=\mathbf{f}^{i}_{x}\cdot \mathbf{f}^{i}_{t}/(||\mathbf{f}^{i}_{x}||||\mathbf{f}^{i}_{t}||)$, $\tau$ is a temperature parameter. The InfoNCE losses, \ie $\mathcal{L}_{t2x}(\mathbf{f}^{i}_{t},\mathbf{f}^{i}_{x},\widetilde{\mathbf{f}}_{x})$ and $\mathcal{L}_{t2z}(\mathbf{f}^{i}_{t},\mathbf{f}^{i}_{z},\widetilde{\mathbf{f}}_{z})$, of language-to-vision can be calculated similarly. Hence, the CMA loss can be formulated as $\mathcal{L}_{cma}=\frac{1}{2}[\mathcal{L}_{x2t}(\cdot)+\mathcal{L}_{z2t}(\cdot)]+\frac{1}{2}[\mathcal{L}_{t2z}(\cdot)+\mathcal{L}_{t2x}(\cdot)]$. 

Intuitively, by optimizing the CMA loss, vision and language embeddings can be well aligned in the feature space as in Fig.~\ref{fig:MMA}. However, the CMA ignores the significant intra-modal supervisory signals (\ie visual template and visual search region) for learning desired multi-modal features. Aligning the visual template with the visual search region enables learning \emph{temporal-invariant features}~\cite{li2018learning,yan2021learning,wang2021transformer}, which are crucial to enhance the discriminative ability of tracking models. To this end, we further propose the IMA to fully utilize the intra-modal temporal supervision information.

\myPara{Intra-modal Alignment.}
As mentioned earlier, IMA aims to learn \emph{temporal-invariant features} within the same modality of positive and negative samples. Since the language prompt mainly contains \emph{global semantic information}, only visual modality is considered in IMA. Specifically, we consider visual search region tokens  $\mathbf{f}_{x}\in \mathbb{R}^{C}$, and visual template tokens $\mathbf{f}_{z}\in \mathbb{R}^{C}$ from the same video as positive pairs, while tokens from different videos as negative pairs. We also apply the contrastive loss to maximize the MI between $\mathbf{f}_{x}$ and $\mathbf{f}_{z}$. Formally, InfoNCE losses between vision tokens can be defined as:
\begin{equation}
    \mathcal{L}_{x2z}(\mathbf{f}^{i}_{x},\mathbf{f}^{i}_{z},\widetilde{\mathbf{f}})\!=\!-\mathbb{E}_{(\mathbf{f}_{x},\mathbf{f}_{z})}\lbrack\log \frac{\exp(sim(\mathbf{f}^{i}_{x},\mathbf{f}^{i}_{z})/\tau)}{\sum_{j=1}^{2(N-1)}\exp(sim(\mathbf{f}^{i}_{x},\widetilde{\mathbf{f}}^{j})/\tau)}\rbrack,
    \label{eq:x2z_loss}
\end{equation}
\begin{equation}
    \mathcal{L}_{z2x}(\mathbf{f}^{i}_{z},\mathbf{f}^{i}_{x},\widetilde{\mathbf{f}})\!=\!-\mathbb{E}_{(\mathbf{f}_{z},\mathbf{f}_{x})}\lbrack\log \frac{\exp(sim(\mathbf{f}^{i}_{z},\mathbf{f}^{i}_{x})/\tau)}{\sum_{j=1}^{2(N-1)}\exp(sim(\mathbf{f}^{i}_{z},\widetilde{\mathbf{f}}^{j})/\tau)}\rbrack,
    \label{eq:z2x_loss}
\end{equation}
where $\mathbf{\widetilde{f}}=\{\widetilde{\mathbf{f}}^{1}_{x},...,\widetilde{\mathbf{f}}^{N-1}_{x},\widetilde{\mathbf{f}}^{1}_{z},...,\widetilde{\mathbf{f}}^{N-1}_{z}\}$ is a set of negative examples for $\mathbf{f}^{i}_{x}$ or $\mathbf{f}^{i}_{z}$, $N$ is the batch size. Then, the IMA loss can be formulated as $\mathcal{L}_{ima} = \frac{1}{2}[\mathcal{L}_{x2z}(\cdot)+\mathcal{L}_{z2x}(\cdot)]$. 

The IMA loss encourages learning representations by aligning temporal-invariant positive pairs within visual modality. Importantly, it enforces the uniformity of vision and language, resulting in a uniform distribution across the whole feature space~\cite{wang2020understanding,yang2022vision}. Therefore, CMA and IMA have complementary advantages in multi-modal learning: on the one hand, CMA encourages matched vision and language embeddings close in the feature space. On the other hand, IMA maximizes the temporal-invariant features between visual tokens, and makes the multi-modal features evenly distributed in the feature space. As shown in Fig.~\ref{fig:MMA}, combining them makes the learned representations more reasonable, and further facilitates joint multi-modal feature learning and interaction.

\subsection{Tracking Head and Loss}
Following~\cite{ye2022joint}, the tracking head is decomposed into two branches of classification and bounding box regression. As shown in Fig.~\ref{fig:framework}, the learned visual search region tokens are first reshaped into a 2D feature map according to the original spatial resolution, followed by a 4-layer fully convolutional network to predict the target classification score map. In the classification branch, a weighted focal loss $\mathcal{L}_{cls}$~\cite{law2018cornernet}, is adopted to enhance the model's ability to distinguish objects from background. The bounding box regression branch is used to predict the center coordinate offset of the object and the size of the object. To regress the center coordinate offset and size of objects, we combine the $\ell_{1}$ loss and the generalized IoU loss $\mathcal{L}_{giou}$~\cite{rezatofighi2019generalized}. The regression loss is calculated as $\mathcal{L}_{reg}=\lambda_{giou}\mathcal{L}_{giou}+\lambda_{{1}}\mathcal{L}_{1}$, where $\lambda_{giou}$ and  $\lambda_{{1}}$are two hyper-parameters.

To train our model in an end-to-end manner, we convert it into a multi-task optimization problem~\cite{zhang2021survey}, simultaneously optimizing classification loss, regression loss, CMA loss, and IMA loss. Finally, the overall loss function for our model is defined as:
\begin{equation}
\mathcal{L}_{total}=\mathcal{L}_{cls}+\mathcal{L}_{reg}+\lambda_{cma}\mathcal{L}_{cma}+\lambda_{ima}\mathcal{L}_{ima},
    \label{eq:total_loss}
\end{equation}
where $\lambda_{cma}$ and $\lambda_{ima}$ are trade-off weights to balance the multi-task optimization problem.

\section{Experiments}

To demonstrate the effectiveness and generalization ability of our approach, we conduct experiments on all five public VL tracking benchmarks to date, including UAV scenes (\ie WebUAV-3M~\cite{zhang2022webuav}), generic scenes (\ie LaSOT~\cite{fan2019lasot}, LaSOT$\rm_{Ext}$~\cite{fan2021lasot}, OTB99-L~\cite{li2017tracking}), and real-synthetic scenes (\ie TNL2K~\cite{wang2021towards}).

\subsection{Implementation Details}

We adopt ViT-Base~\cite{dosovitskiy2020image} as the architecture of the All-in-One transformer backbone. It is stacked by $L$ (\ie 12) transformer encoder layers, and each layer contains two sub-layers, \ie a multi-head self-attention layer and a feed-forward network. Each sub-layer is a residual connection structure, followed by a layer normalization. To accelerate convergence, we initialize our backbone with MAE-pretrained weights~\cite{He2022MAE}. The visual template and visual search region are $2^{2}$ times and $4^{2}$ times of the target bounding box, and then resized to $128\times128$ and $256\times256$, respectively. We use bert-base-uncased tokenizer~\cite{devlin2018bert} to tokenize language prompts.

Our method is trained on an Ubuntu server with two NVIDIA RTX 3090 GPUs. The training data includes training splits of LaSOT~\cite{fan2019lasot}, GOT-10k~\cite{huang2019got}, TrackingNet~\cite{muller2018trackingnet}, COCO~\cite{lin2014microsoft}, OTB99-L~\cite{li2017tracking}, TNL2K~\cite{wang2021towards}, WebUAV-3M~\cite{zhang2022webuav}, and VisualGenome~\cite{krishna2017visual}. For several datasets (\ie ~\cite{huang2019got,muller2018trackingnet,lin2014microsoft}) without natural language prompts, we use class names as pseudo language labels similar to~\cite{guo2022divert}. The tracker is optimized using an AadmW optimizer~\cite{loshchilov2017decoupled} with an initial learning rate $4\times10^{-4}$. The total epoch is 300. The weight decay factor is $1\times10^{-4}$ after 240 epochs. Following~\cite{ye2022joint}, the hyper-parameters $\lambda_{{giou}}$ and $\lambda_{1}$ are set to 2 and 5. While $\lambda_{cma}$ and $\lambda_{ima}$ are set to 1 and 1 without parameter optimization. We set the temperature parameter $\tau=0.5$. The batch size $N$ is 32. Following~\cite{wu2015otb,zhang2022webuav}, we adopt the one-pass evaluation with five metrics, \ie precision ($P$), normalized precision ($P_{norm}$), success rate (AUC), complete success rate (cAUC), and accuracy (ACC) to measure the tracking performance. % Following~\cite{zhou2023joint}, we test the proposed method on a single NVIDIA 3090 GPU. 

\begin{table}[t]
	\caption{Ablation study of our approach on the LaSOT dataset. We retrain~\cite{ye2022joint} as our baseline. ``AOT-[CLS]'' and ``AOT-[Mean]'' represent All-in-One transformer (AOT) using [CLS] token and mean token of language prompt, respectively.}
 \vspace{-0.3cm}
	\label{tab:AblationStudy}
	\setlength{\tabcolsep}{3pt}
	\begin{center}
		\begin{tabular}{l|ccc}
		%\toprule
		\hline
		Method  & AUC (\%) & $P_{norm}$ (\%) & $P$ (\%)\\
		\hline
            Baseline & 62.2 & 70.5 & 66.3 \\
            \hline
            Baseline w/ AOT-[CLS] & 63.9 & 72.4 & 67.9 \\
            
            Baseline w/ AOT-[Mean] & 64.0 & 72.6 & 68.4 \\

            \hline
            Baseline w/ AOT-[Mean] and CMA & 64.1 &  72.7  & 68.6\\
            
            Baseline w/ AOT-[Mean] and MMA & \textbf{64.4} & \textbf{72.8} & \textbf{68.8} \\
		\hline
		%\bottomrule
		\end{tabular}
	\end{center}
\end{table}

\begin{figure}[t]
  \centering
  \includegraphics[width=\linewidth]{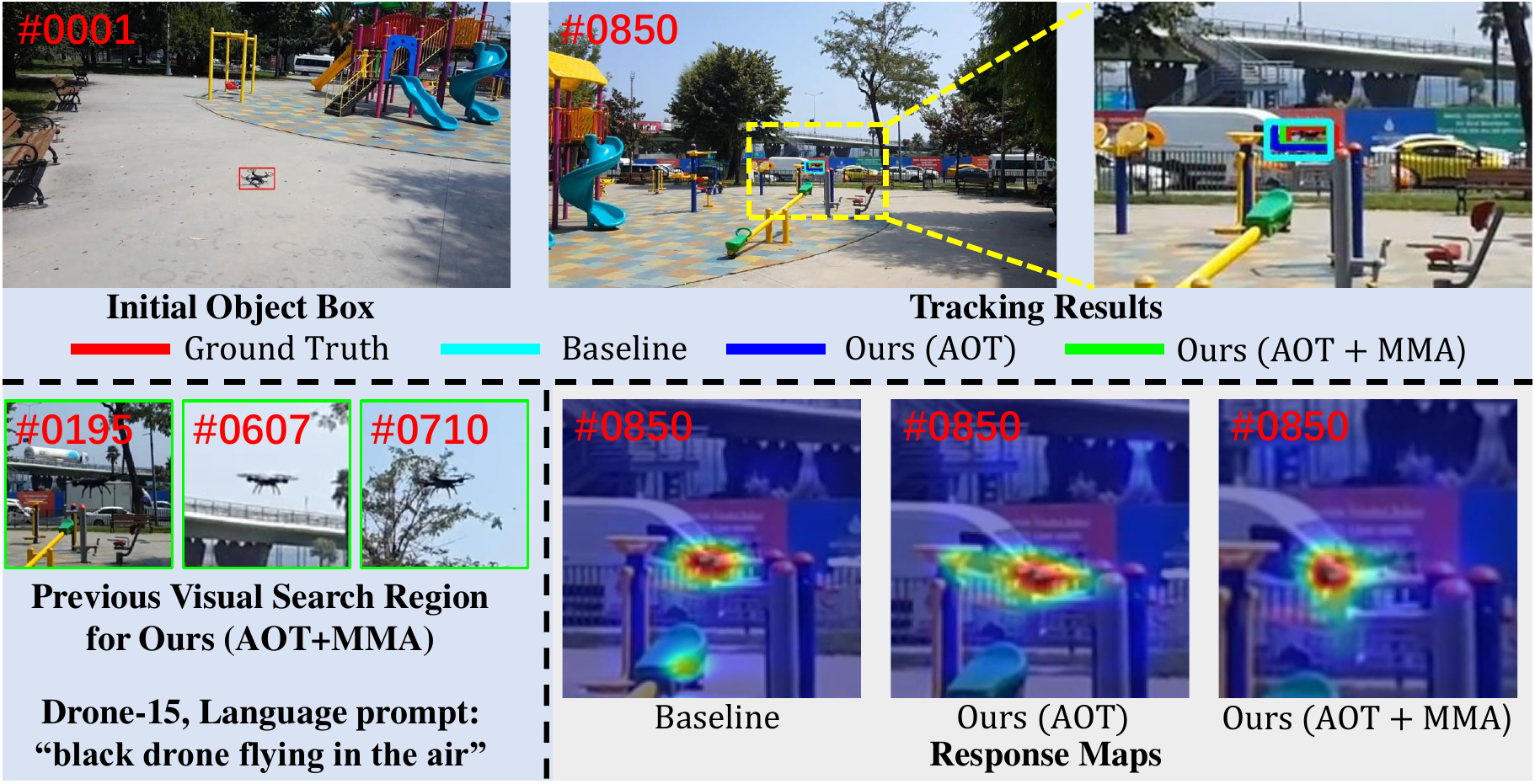}
  \vspace{-0.5cm}
  \caption{Visualization for revealing the target-aware capability of the All-in-One framework. ``AOT'' denotes our approach only with All-in-One transformer, ``AOT+MMA'' denotes our approach with both All-in-One transformer and multi-modal alignment module.}
  \label{fig:visualization}
  \vspace{-0.3cm}
\end{figure}

\subsection{Ablation Study and Analysis}
We first conduct ablation experiments trained on the LaSOT training set and evaluated on the LaSOT test set to validate diﬀerent components of our approach.

%\myPara{Training Loss \& Accuracy.} Two training curves: $\mathcal{L}_{reg}$ loss vs iteration, IoU score vs iteration show convergence speed.

\myPara{Impact of All-in-One Transformer  (AOT).} To analyze the impact of AOT, we train two trackers, \ie AOT-[CLS] and AOT-[Mean] using [CLS] token and mean token of language prompt in AOT. From Tab.~\ref{tab:AblationStudy}, we can find that the AOT obviously boosts the tracking performance. Specifically, the $P$ scores are improved by $1.6\%$ (from $66.3\%$ to $67.9\%$) and $2.1\%$ (from $66.3\%$ to $68.4\%$), respectively compared with the baseline method. Importantly, using the mean token is slightly better than the [CLS] token. We speculate that the possible reason is that the mean token can provide more semantic information about the target compared with the [CLS] token. Therefore, the mean token is our default setting in AOT.

\myPara{Impact of Cross-modal Alignment (CMA).} From Tab.~\ref{tab:AblationStudy}, we can see the CMA component improves tracking performance by $0.1\%$, $0.1\%$, and $0.2\%$ in terms of AUC, $P_{norm}$, and $P$ scores, respectively. This validates the CMA is beneficial to align vision and language embeddings in the feature space and improve tracking accuracy.

\myPara{Impact of Intra-modal Alignment (IMA).} By combining the CMA and IMA, we improve the tracking AUC by $0.4\%$ (from $64.0\%$ to $64.4\%$), $P_{norm}$ by $0.2\%$ (from $72.6\%$ to $72.8\%$), and $P$ by $0.4\%$ (from $68.4\%$ to $68.8\%$), as shown in Tab.~\ref{tab:AblationStudy}. The significant performance gains demonstrate that the MMA module makes the distributions of vision and language embeddings more reasonable in the feature space, and facilitates feature learning and interaction.

\begin{table}[t]
	\caption{Impact of language prompts (\ie sentence and class). ``Ours-S'' and ``Ours-C'' denote our approaches (w/ AOT and MMA) training with sentence and class prompts, respectively.}
 \vspace{-0.3cm}
	\label{tab:language_prompts}
	\setlength{\tabcolsep}{4.4pt}
	\begin{center}
		\begin{tabular}{c|c|c|ccc}
		%\toprule
		\hline
	Method &	Training  & Test & AUC (\%) & $P_{norm}$ (\%) & $P$ (\%)\\
		\hline  
            %% 1/1000
         \multirow{2}{*}{Ours-S} &  Sentence &  Class & 63.4 & 72.0 & 67.8 \\
          &  \textbf{Sentence} &  \textbf{Sentence} & \textbf{64.4} & \textbf{72.8} & \textbf{68.8} \\
            \hline
        \multirow{2}{*}{Ours-C} &   Class &  Sentence & 63.2 & 71.9 & 67.4 \\
         &   \textbf{Class} &  \textbf{Class} & \textbf{64.6} & \textbf{73.2} & \textbf{68.9} \\
   
            %% 1/10000
            % Sentence &  Sentence & 64.37 & 72.80 & 68.80 \\
            % Sentence &  Class & 63.38 & 72.00 & 67.78 \\
            % Class &  Class & 64.64 & 73.22 & 68.85 \\
            % Class &  Sentence & 63.23 & 71.86 & 67.40 \\
		\hline
		%\bottomrule
		\end{tabular}
	\end{center}
\end{table}

\begin{figure}[t]
  \centering
  \includegraphics[width=\linewidth]{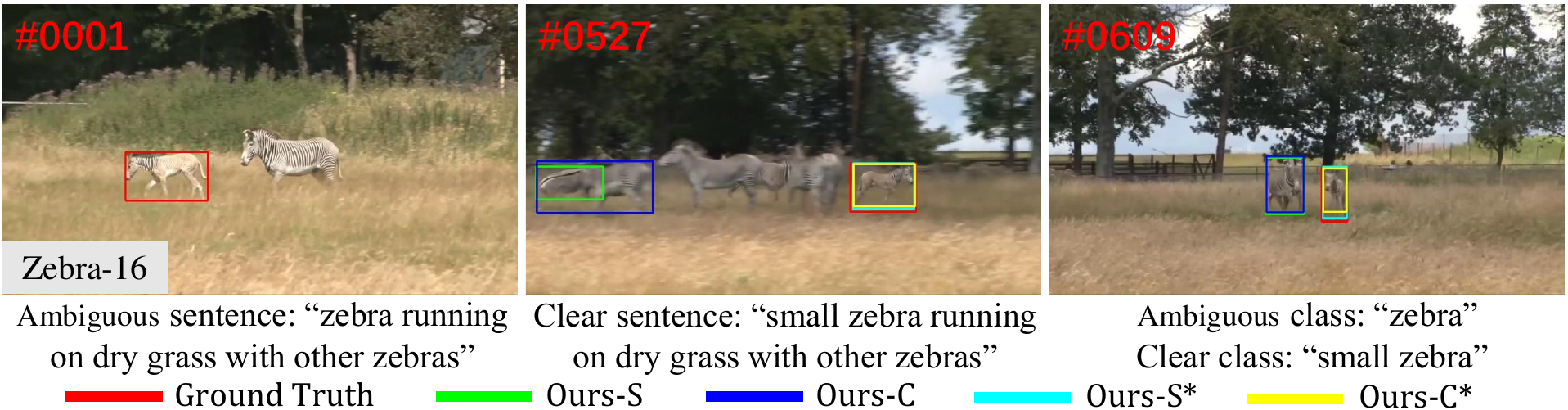}
  \vspace{-0.5cm}
  \caption{Analysis of the effect of ambiguous language prompts on the LaSOT test set. $^{*}$ indicates that our approach is tested with a clear sentence prompt or a clear class prompt.}
  \label{fig:language_prompt}
  \vspace{-0.3cm}
\end{figure}

\begin{table*}[t]
\caption{
Performance comparison on four benchmarks, including LaSOT, LaSOT$_{\rm Ext}$, OTB99-L, and TNL2K. We compare All-in-One with convolutional neural network (CNN)-based, CNN-VL, transformer (Trans)-based, and Trans-VL methods.}
\vspace{-0.3cm}
\label{table:four_datasets}
\begin{threeparttable}
\setlength{\tabcolsep}{1.61mm}{
\begin{tabular}{c|c|ccc|cc|cc|cc|c}
%\toprule
\hline
    \multicolumn{1}{c|}{\multirow{2}[1]{*}{Type}} & 
    \multicolumn{1}{c|}{\multirow{2}{*}{Method}} &           \multicolumn{3}{c|}{{LaSOT }} & \multicolumn{2}{c|}{{LaSOT$_{\rm Ext}$ }} &
	\multicolumn{2}{c|}{OTB99-L} & 
	\multicolumn{2}{c|}{TNL2K} & 
	\multirow{2}{*}{Speed (FPS)} \\
			 
    \cline{3-5} \cline{6-7} \cline{8-9} \cline{10-11}
	\multicolumn{1}{c|}{} & \multicolumn{1}{c|}{}  & AUC (\%) &  $P$ (\%) & $P_{ norm}$ (\%) &  AUC (\%) & $P$ (\%) &  AUC (\%)  & $P$ (\%) & AUC (\%)& $ P$ (\%) & \\
\hline

\multirow{11}{*}{CNN-based} & SiamRCNN~\cite{voigtlaender2020siam}  & 64.8  & 68.4 & 72.2 & - & - & 70.0 & 89.4 & 52.3 & 52.8 & 4.7  \\

 & PrDiMP~\cite{danelljan2020probabilistic}  & 59.8  & 60.8 & 68.4 & - & - & 69.5 & 89.5 & 47.0 & 45.9 & 30.0  \\

 & AutoMatch~\cite{zhang2021learn}   & 58.3  & 59.9 & - & 37.6 & 43.0 & 71.6  & 93.2 & 47.2 & 43.5 & 50.0  \\

 & Ocean~\cite{zhang2020ocean}    & 56.0 & 56.6 & 65.1  & - &  - & 68.0 & 92.1 &  38.4 & 37.7 & 58.0 \\

 & ATOM~\cite{danelljan2019atom}  & 51.5 & 50.5 & 57.6 & 37.6 & 43.0  & 67.6 & 82.4 & 40.1 & 39.2 & 30.0  \\

 & SiamRPN++~\cite{li2019siamrpn++}  & 49.6 & 49.1 & 56.9 & 34.0 & 39.6 & 63.8 & 82.6 & 41.3 & 41.2 & 35.0  \\

 & GlobalTrack~\cite{huang2020globaltrack} & 51.7 & 52.8 & 59.7 &  35.6 & 41.1 & - & - & 40.5 & 38.6  & 6.0  \\

 & SiamFC~\cite{bertinetto2016fully}  & 33.6 & 33.9 & 42.0 & 23.0 & 26.9 & 58.7 & 79.2 & 29.5 & 28.6 & 86.0  \\

 & SiamCAR~\cite{guo2020siamcar} & 50.7 & 51.0 & 60.0 & 33.9 &  41.0 & 68.8 & 89.1 & 35.3 & 38.4  & 52.3  \\

\hline

\multirow{2}{*}{CNN-VL} & SNLT~\cite{feng2021siamese}  & 54.0 & 57.6 & 63.6 & 26.2 & 30.0 & 66.6 & 80.4 & 27.6 & 41.9 & 50.0  \\

 & VLT$_{\rm SCAR}$~\cite{guo2022divert}$^{*}$  & 63.9 & 67.9 & 73.3 & 44.7 & 51.6 & 73.9 &  89.8 & 49.8 & 51.0 & 43.0 \\
\hline

\multirow{4}{*}{Trans-based} & STARK-ST50~\cite{yan2021learning}  & 66.4 & 71.2 & 76.3 & 47.8 & 55.1 & 69.6 & 91.4 & - & -  & 40.0 \\

 & TrDiMP~\cite{wang2021transformer}  & 63.9 & 66.3 & - & - & -  & 70.5 & 92.5 & - & - & 26.0  \\

 & TransT~\cite{chen2021transformer}  & 64.9 & 69.0 & 73.8 & 44.8 & 52.5 & 70.8 & 91.2 & 50.7 & 51.7 & 50.0  \\

  & OSTrack~\cite{ye2022joint}  & 69.1 & 75.2 & 78.7 & 47.4 & 53.3 & 70.6 & 92.1 & 54.3  & 56.3  & {105.4}  \\
 
\hline

%\multirow{3}{*}{Trans-VL} & JointNLT~\cite{zhou2023joint}  & 56.9  & 59.3 & - & - & - & 59.2 & 77.6 & 54.6 & 55.0  & 39.0  \\  % NL version

%\multirow{3}{*}{Trans-VL} & JointNLT~\cite{zhou2023joint}  & 60.4  & 63.6 & - & - & - & 65.3 & 85.6 & \textbf{56.9} & \textbf{58.1}  & 39.0  \\  % NL + Box version

\multirow{2}{*}{Trans-VL}  & VLT$_{\rm TT}$~\cite{guo2022divert}$^{*}$  & 67.3  & 72.1 & 77.6 & 48.4 & 55.9  & \textbf{76.4} & \textbf{93.1} & 53.1 & 53.3 & 35.0  \\

& \textbf{All-in-One (Ours)}  & \textbf{71.7}  & \textbf{78.5} & \textbf{82.4} & \textbf{54.5} & \textbf{62.0} & {71.0} & {93.0} & \textbf{55.3} & \textbf{57.2}  & \textbf{60.0}  \\

\hline
%\bottomrule
\end{tabular}
}
 \begin{tablenotes}
        \footnotesize
        \item[*] For this tracker~\cite{guo2022divert}, results were obtained from four different models. The best result is reported on each dataset.
      \end{tablenotes}
    \end{threeparttable}
\vspace{-0.1cm}
\end{table*}

\myPara{Visualization.}
To further investigate the effectiveness of our All-in-One framework, we visualize the response maps and the tracking results in Fig.~\ref{fig:visualization}. With the AOT, our approach highlights the target region due to the language prompt, even with complex background distractions. Combining AOT and MMA, our approach has a more unambiguous and discriminative response, and predicts a more precise bounding box. Previous visual search regions also demonstrate that our approach can focus on the real target when facing some complex scenarios, such as occlusion and background clutter.

\myPara{Sentence Prompts \emph{vs.} Class Prompts.} To analyze the impact of language prompts, we train two trackers, \ie Ours-S and Ours-C with sentence (original language prompt) and class (class name of the video) prompts on the LaSOT training set. From Tab.~\ref{tab:language_prompts}, we have some observations upon inspection. First, better tracking results are achieved when the language prompts for training and testing are consistent, \ie training using sentences/classes and testing using sentences/classes. Second, the best performance ($64.6\%$ in AUC, $73.2\%$ in $P_{norm}$, $68.9\%$ in $P$) is obtained using class prompts training and class prompts testing on the LaSOT dataset. We speculate that trackers are sensitive to ambiguous language prompts. Compared with sentence prompts, class prompts may bring about less ambiguity for both training and evaluation~\cite{zhou2023joint,zhang2022webuav}. Additionally, as shown in Fig.~\ref{fig:language_prompt}, given ambiguous language prompts our trackers fail to localize the real object. A potential solution is to provide clear sentence prompts or clear class prompts (see Fig.~\ref{fig:language_prompt}), both of which enable our trackers to accurately localize the real object.

\myPara{Speed Analysis.} Real-time tracking is urgently demanded in many practical applications~\cite{zhang2020accurate,wang2021towards}. Our one-stream framework achieves joint multi-modal feature extraction and interaction, and has great efficiency. Tab.~\ref{table:four_datasets} shows that the average inference speed of our approach is around 60 frames per second (FPS) without model acceleration. The speed of our approach is obviously faster than that of many SOTA real-time trackers~\cite{li2019siamrpn++,zhang2021learn} and common video flow~\cite{tekalp2015digital}, demonstrating that applying it to real-world applications is imperceptible regarding time consumption.

\begin{figure}[t]
\vspace{-0.3cm}
\centering
\subfloat{\includegraphics[width =0.5\columnwidth]{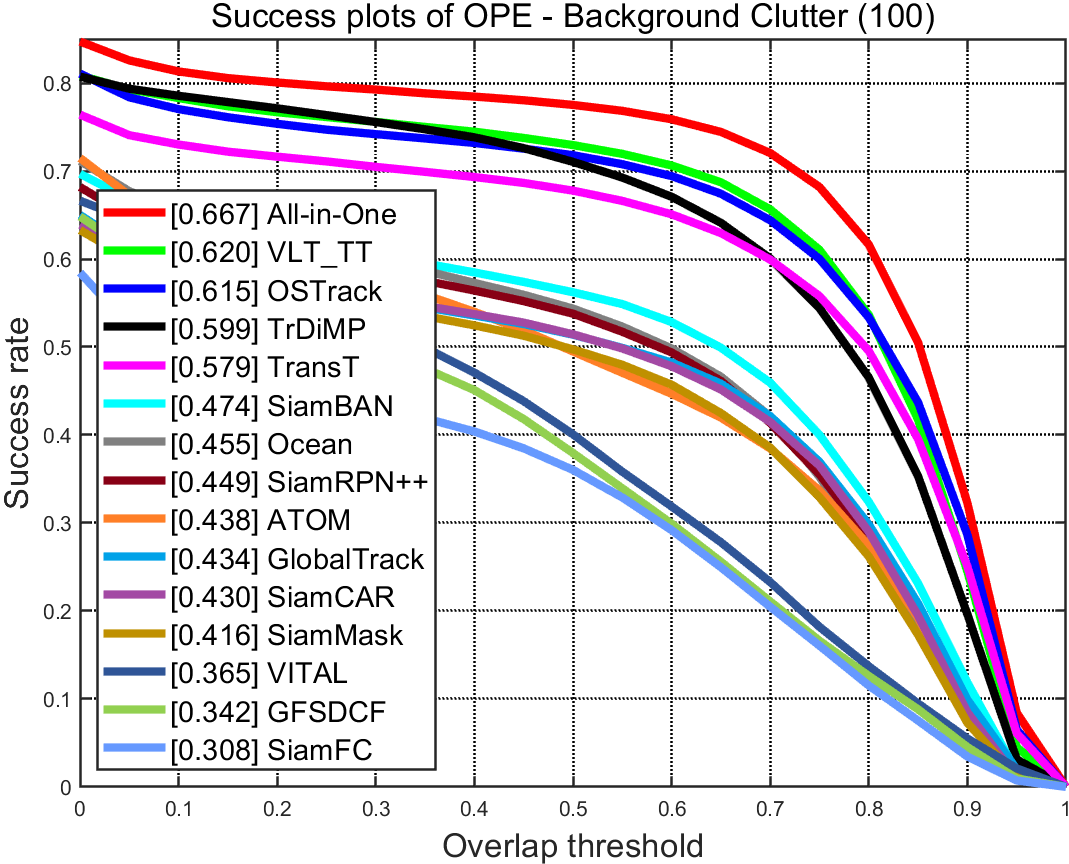}}
~
\subfloat{\includegraphics[width =0.5\columnwidth]{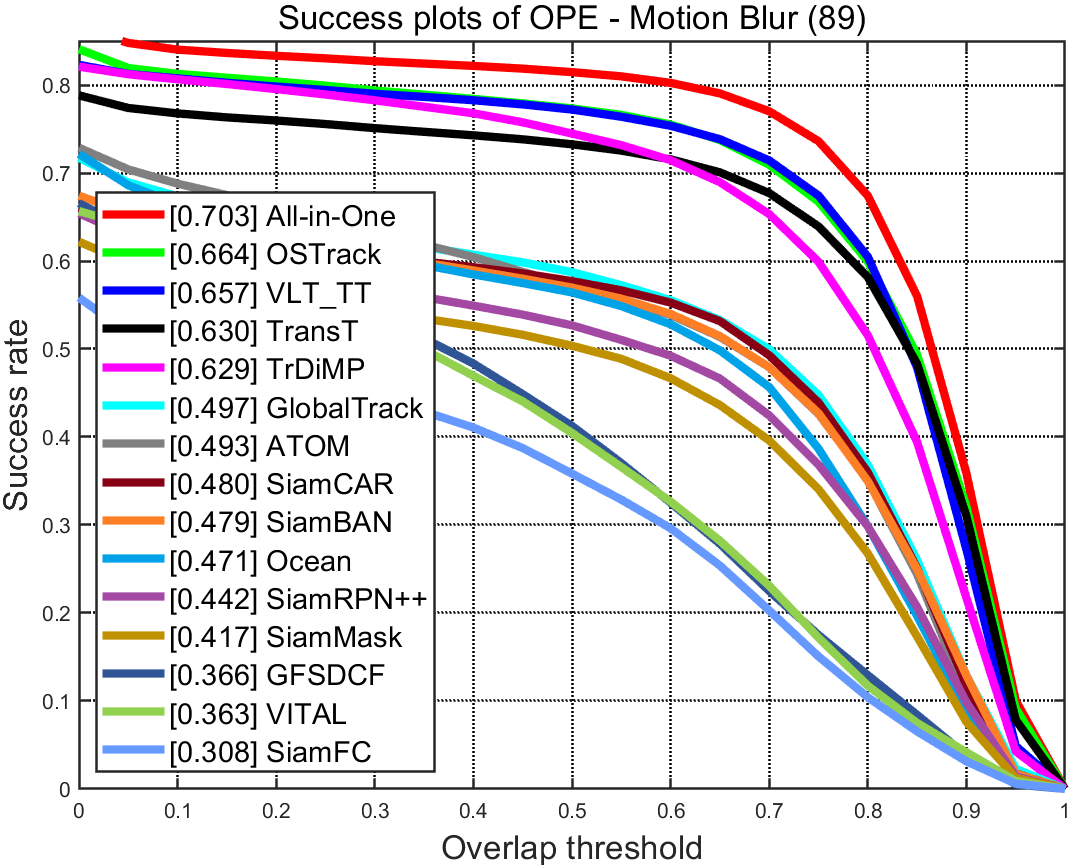}}
%~
%\subfloat{\includegraphics[width =0.7\columnwidth]{figs/LaSOT/IV_overlap_OPE_AUC.png}}
%\\
%\vspace{-0.2cm}
%\subfloat{\includegraphics[width =0.71\columnwidth]{figs/LaSOT/FM_overlap_OPE_AUC.png}}
%~
%\subfloat{\includegraphics[width =0.71\columnwidth]{figs/LaSOT/FOC_overlap_OPE_AUC.png}}
%~
%\subfloat{\includegraphics[width =0.71\columnwidth]{figs/LaSOT/DEF_overlap_OPE_AUC.png}}
\vspace{-0.3cm}
\caption{Performance of trackers on two challenging attributes on the LaSOT test set.}
%\caption{Attribute-based evaluation on the LaSOT test set. AUC score is used to rank diﬀerent trackers.}
\label{fig:LaSOT_attribute}
\vspace{-0.3cm}
\end{figure}

\subsection{Evaluation on Generic Scenes}
\myPara{LaSOT.} LaSOT~\cite{fan2019lasot} is a densely annotated large-scale VL tracking dataset that contains 1,120 videos for training and 280 long-term videos for evaluation. In this dataset, objects disappear and reappear frequently, making long-term tracking in generic scenes highly challenging. From Tab.~\ref{table:four_datasets}, we can observe that our approach sets a new SOTA on LaSOT, which provides compelling evidence for long-term tracking and suggests that our approach is capable of recognizing objects in extremely long videos. Fig.~\ref{fig:LaSOT_attribute} demonstrates All-in-One outperforms other trackers on two challenging attributes, \ie background clutter and motion blur.

\myPara{LaSOT$_{\rm Ext}$.} LaSOT$_{\rm Ext}$~\cite{fan2021lasot} is the extended version of~\cite{fan2019lasot}, which comprises 150 manually annotated videos. Tab.~\ref{tab:AblationStudy} indicates All-in-One surpasses all previous advanced trackers and obtains the best AUC score of $54.5\%$, gaining a significant improvement of $7.1\%$ compared with the current SOTA one-stream tracker OSTrack~\cite{ye2022joint}.

\myPara{OTB99-L.} OTB99-L~\cite{li2017tracking} is an early VL tracking dataset contains 51 videos for training and 48 videos for public evaluation. As shown in Tab.~\ref{table:four_datasets}, compared with recent SOTA trackers, our tracker achieves comparable results, which validates the effectiveness of our tracker.

\subsection{Evaluation on Real-Synthetic Scenes}
\myPara{TNL2K.} TNL2K~\cite{wang2021towards} is a recently released dataset, which comprises 1,300 videos for training and 700 videos for evaluation in real and synthetic (\eg cartoon videos and virtual game videos) scenes with diverse challenging factors, such as significant appearance variation and adversarial samples. Results in Tab.~\ref{table:four_datasets} show that our approach achieves the highest AUC ($55.3\%$) and $P$ (57.2\%) scores, demonstrating the generalization ability of All-in-One.

\begin{table}[t]
	\caption{Comparison of our approach with SOTA trackers on the WebUAV-3M test set using ACC score.}
 \vspace{-0.4cm}
	\label{tab:webuav3m_accuracy}
	\setlength{\tabcolsep}{5.8pt}
	\begin{center}
		\begin{tabular}{cc|cc}
		%\toprule
		\hline
		Method  & ACC (\%) & Method &  ACC (\%) \\
		\hline
            SiamFC~\cite{bertinetto2016fully} &  34.9 & SiamBAN~\cite{ChenZLZJ20} & 44.9 \\
            
             Ocean~\cite{zhang2020ocean}  &  37.0 & VLT$_{\rm SCAR}$~\cite{guo2022divert} &  45.3  \\
             
            DiMP~\cite{BhatDGT19iccv} &  37.5 & AutoMatch~\cite{zhang2021learn}  & 46.1  \\
            
            TrDiMP~\cite{wang2021transformer} &  41.2 & TransT~\cite{chen2021transformer} & 46.6 \\
            
            SiamCAR~\cite{guo2020siamcar}  & 42.2  & VLT$_{\rm TT}$~\cite{guo2022divert} & 47.5 \\
            
            SiamRPN++~\cite{li2019siamrpn++}  &  44.2 & \textbf{All-in-One (Ours)} & \textbf{57.6} \\
    
		\hline
		%\bottomrule
		\end{tabular}
	\end{center}
 \vspace{-0.3cm}
\end{table}

\subsection{Evaluation on UAV Scenes}
\begin{figure}[t]
\vspace{-0.5cm}
\centering
\subfloat{\includegraphics[width =0.5\columnwidth]{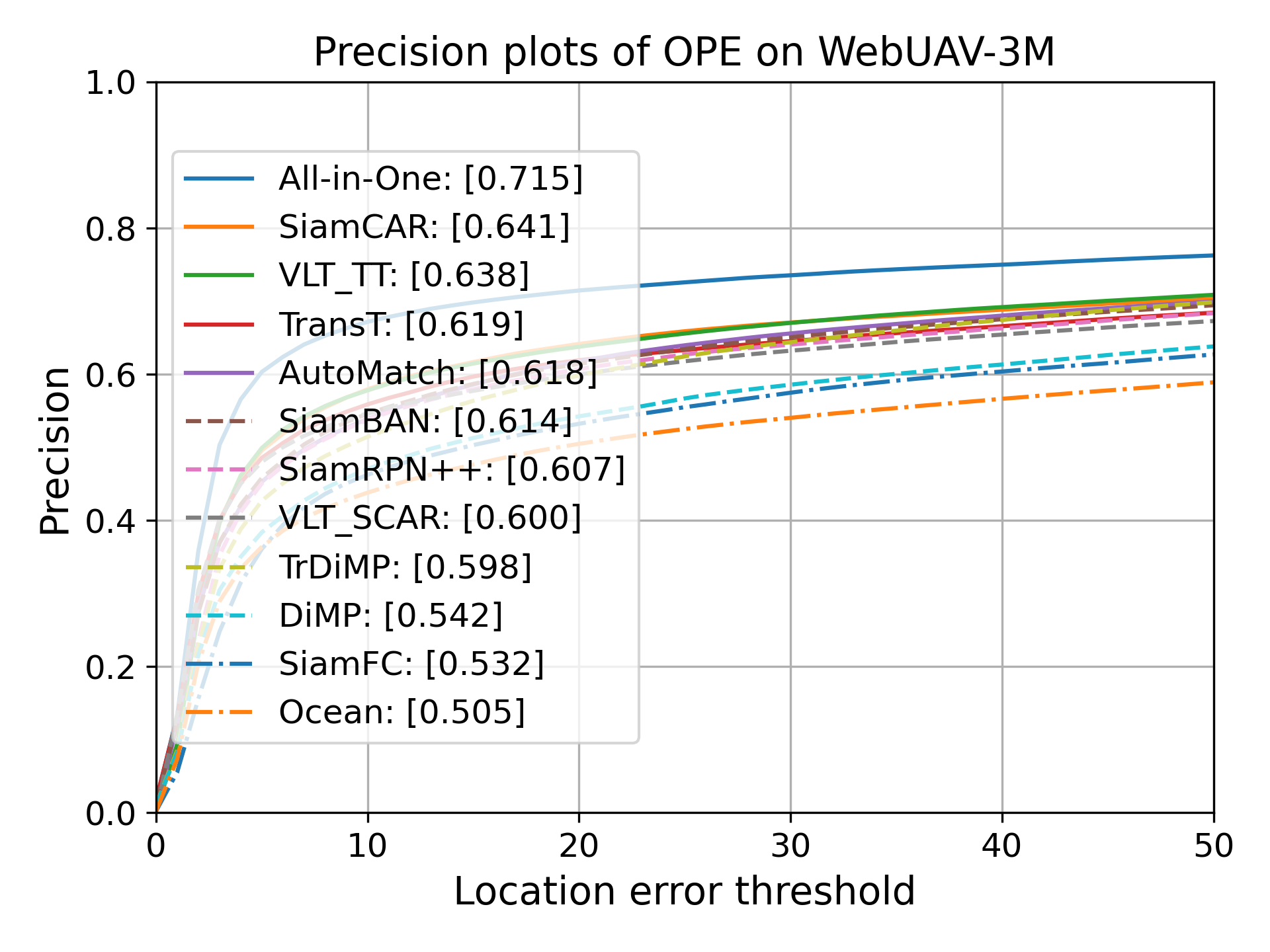}}
~\subfloat{\includegraphics[width =0.5\columnwidth]{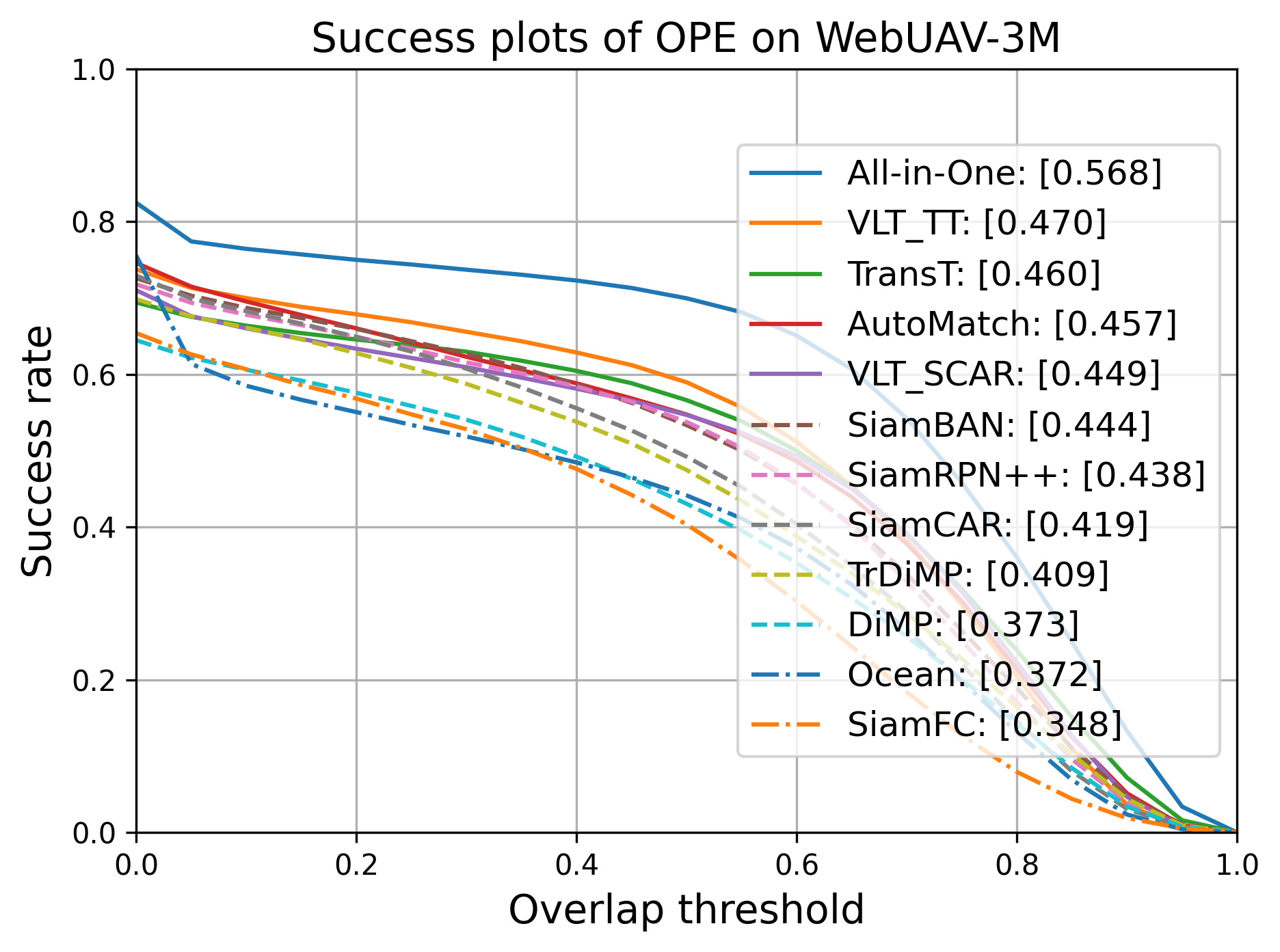}}
\\
\vspace{-0.4cm}
\subfloat{\includegraphics[width =0.5\columnwidth]{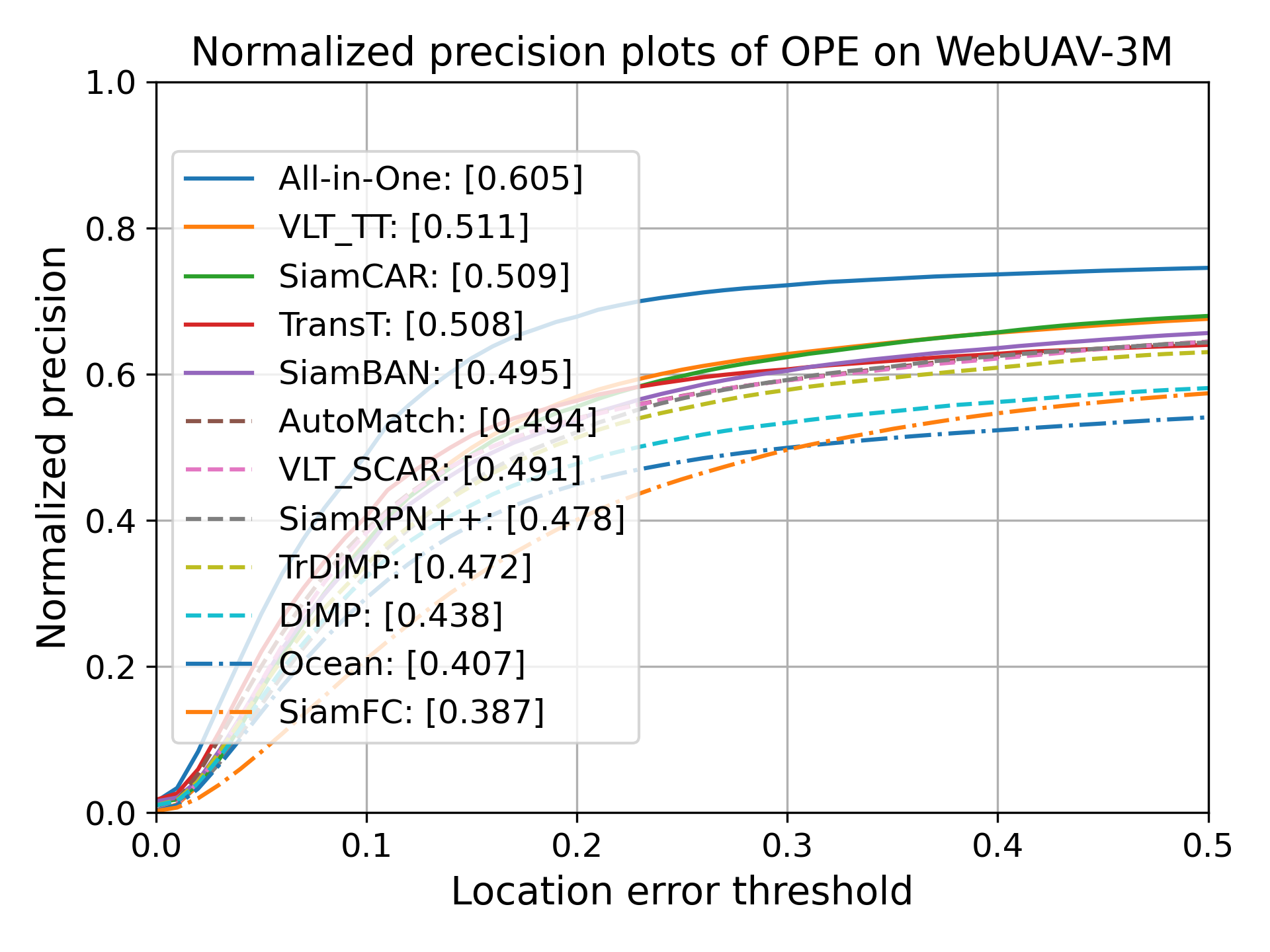}}
~\subfloat{\includegraphics[width =0.5\columnwidth]{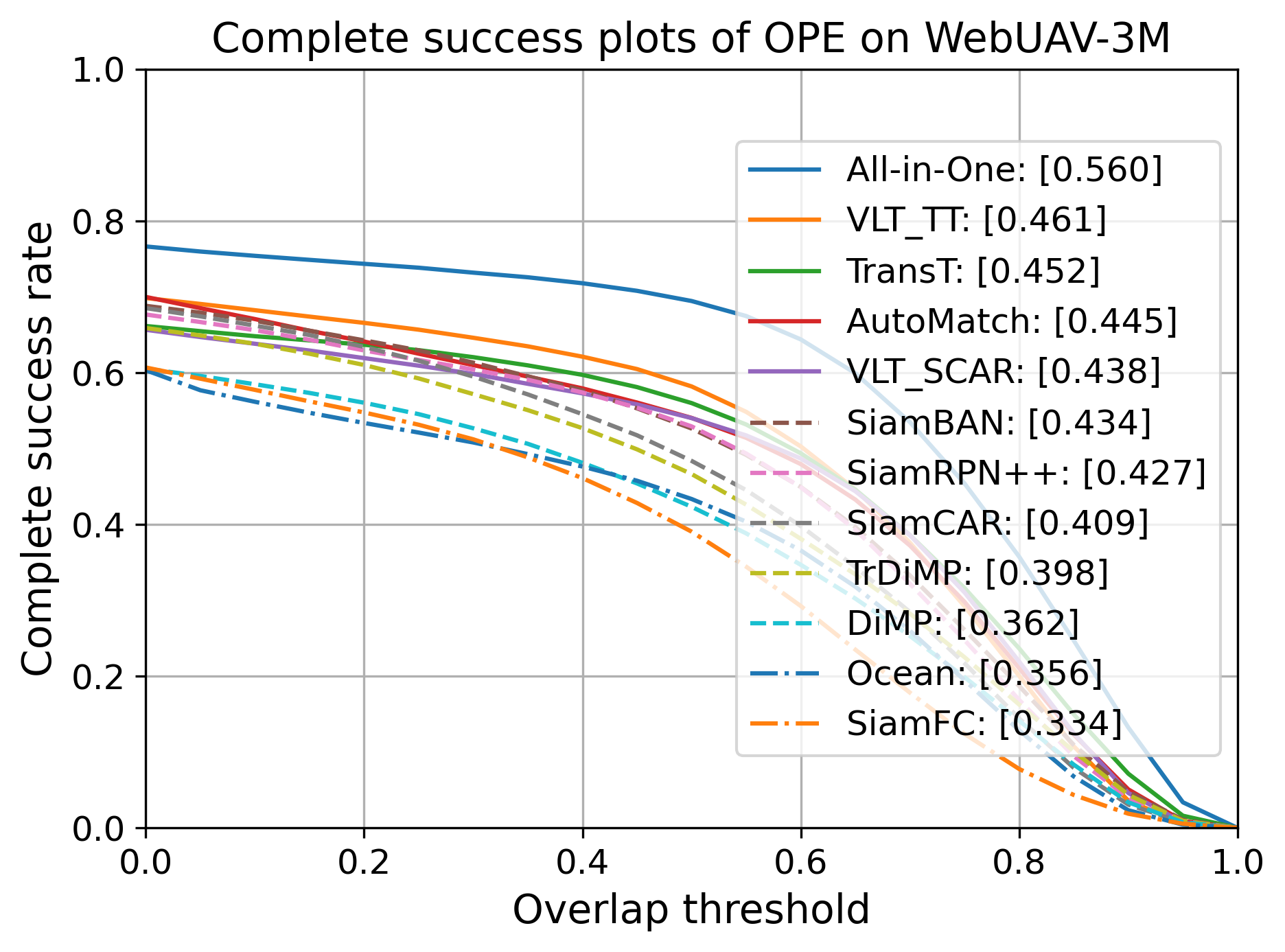}}
\vspace{-0.4cm}
\caption{Evaluation on the WebUAV-3M test set. The scores of $P$/AUC/$P_{norm}$/cAUC are presented in the legend.}
\label{fig:WebUAV-3M_test_set}
\vspace{-0.3cm}
\end{figure}

\myPara{WebUAV-3M.} WebUAV-3M~\cite{zhang2022webuav} is the latest million-scale UAV tracking dataset with visual bounding box, language and audio annotations, which contains 4,500 videos and offers over 200 target categories. UAV tracking scenes are extremely challenging due to continuous viewpoints changes, motion blurs, low resolutions, \etc. As reported in Tab.~\ref{tab:webuav3m_accuracy}, All-in-One outperforms other visual trackers and VL trackers in tracking accuracy. Furthermore, our tracker improves $P$/AUC/$P_{norm}$/cAUC by a large margin as shown in Fig.~\ref{fig:WebUAV-3M_test_set}. Notably, with a simple and general unified model architecture, our tracker outperforms the most competitive VL tracker VLT$_{\rm TT}$~\cite{guo2022divert} by $7.7\%$ in $P$, $9.5\%$ in $P_{norm}$, $9.8\%$ in AUC, and $9.9\%$ in cAUC.

\subsection{Qualitative Performance}
\begin{figure}[t]
  \centering
\includegraphics[width=1.0\linewidth]{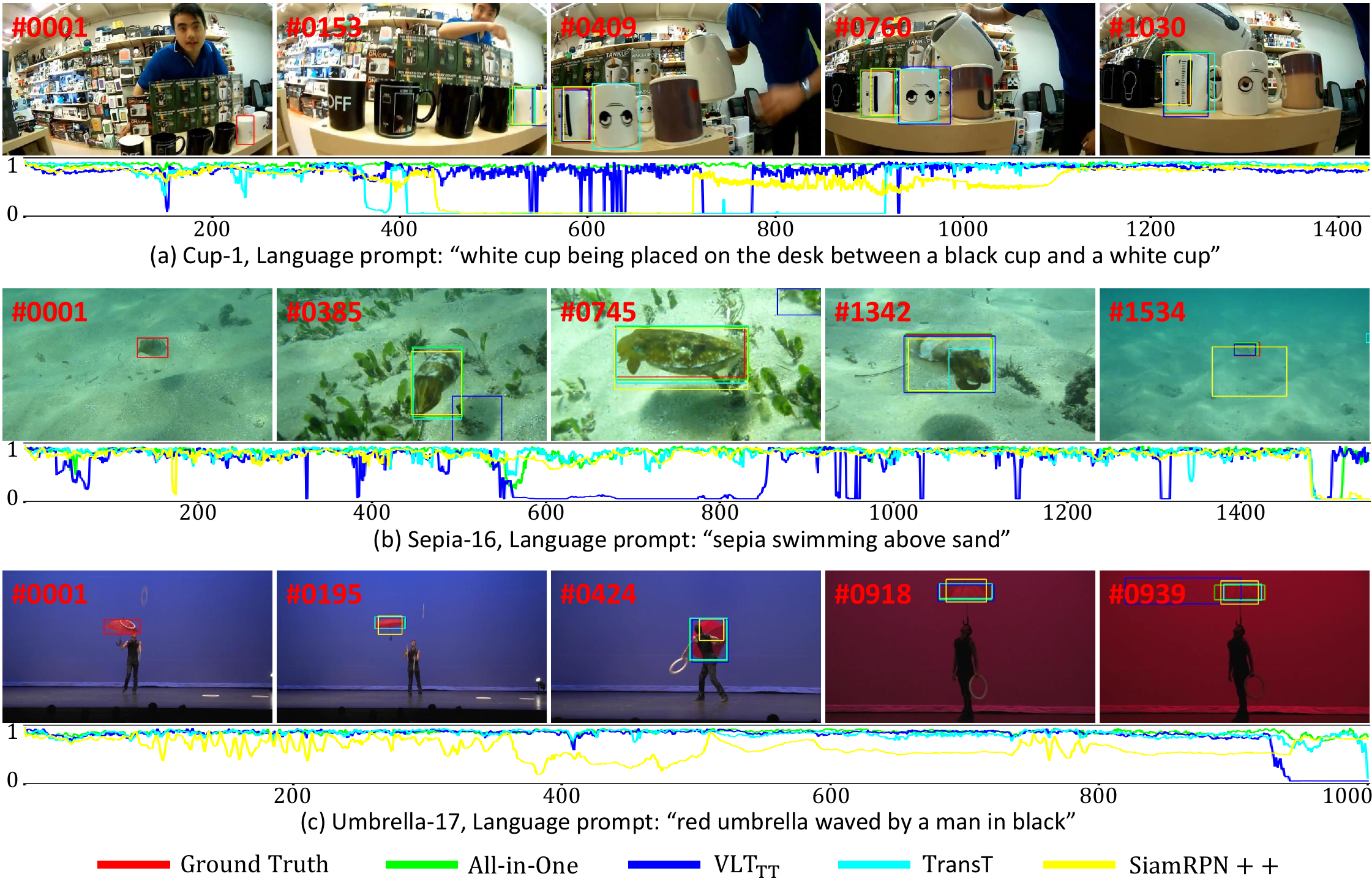}
\vspace{-0.65cm}
  \caption{Qualitative comparison on three challenging video sequences from the LaSOT test set. We present some representative tracking results (top), ACC score (middle) and language prompt (bottom) for each video.}
  \vspace{-0.35cm}
   \label{fig:qualitative_performance}
\end{figure}

As shown in Fig.~\ref{fig:qualitative_performance}, we compare All-in-One with three SOTA trackers (\ie VLT$_{\rm TT}$~\cite{guo2022divert}, TransT~\cite{chen2021transformer}, and SiamRPN++~\cite{li2019siamrpn++}) on three videos from the LaSOT test set, in which the main challenges include similar distractors, severe viewpoint changes, background clutter, appearance variation, occlusion, and extreme illumination. We can see that All-in-One is more robust than other methods. For instance, the prior most competitive VL tracker VLT$_{\rm TT}$ gradually loses the target as the target appearance varies in the video of \emph{Sepia-16} (the second video in Fig.~\ref{fig:qualitative_performance}). By contrast, All-in-One provides accurate and stable prediction results, demonstrating the effectiveness of our unified framework in complex environments.

\section{Conclusion and Discussion}
\vspace{-0.1cm}
\myPara{Conclusion.}
In this work, we present All-in-One, a new framework for multi-modal VL tracking, which contains the All-in-One transformer and the MMA module. The core insight is to is establish bidirectional information ﬂow between well-aligned visual and language signals as early as possible via a unified transformer backbone. Besides, the MMA module based on cross-modal and intra-modal contrastive objectives enables to learn more reasonable VL representations, which effectively facilitates joint multi-modal feature learning and interaction. Extensive experiments on multiple VL tracking benchmarks have demonstrated the effectiveness and generalization of our approach against SOTA trackers. 

\vspace{-0.05cm}
\myPara{Discussion.} We first provide a discussion to demonstrate that developing a foundation model, \eg All-in-One, for VL tracking is valuable in the era of large language/vision models. (1) As the echoes of large language models (\eg ChatGPT~\cite{ChatGPT}, GPT-4~\cite{GPT4}) remarkable success continue to permeate the natural language community, its formidable successors, \eg ViT-22B~\cite{dehghani2023scaling}, have emerged in the computer vision community. Although they have emergent abilities~\cite{ChatGPT}, the huge training cost (\eg thousands of GPUs) and terrible environment unfriendliness cannot be ignored~\cite{koubaa2023gpt}. Instead, we believe that training a foundation model for a specific task is more flexible and affordable for research purposes. (2) Despite the breakthroughs in multi-modal models~\cite{wang2022image,alayrac2022flamingo,chen2022pali,Girdhar_2023_CVPR}, they have not achieved the same success as large language models, highlighting the need to explore foundation models in the multi-modal domain. All-in-One is designed to be such a foundation model for multi-modal VL tracking. (3) All-in-One has great potential to become a foundation model for multi-modal tracking because it can enable more accurate and efficient processing of multi-modal data, which fully utilizes both vision and language information. Our model not only learns all modalities in one backbone (All-in-One), but trains once and generalizes well to all VL tracking datasets (Once-for-All) with complex and user-defined language prompts. (4) Additionally, having a streamlined and standardized foundation model for multi-modal tracking can facilitate the development of more complex and specialized models in the future, allowing for even more sophisticated analysis and understanding of multi-modal data.

Our work still has the following two limitations. (1) Our approach is designed to localize objects of interest based on object boxes and language prompts. Inevitably, it suffers from inaccurate language prompts, such as ambiguous language descriptions, and the states (\eg position and appearance) of objects change significantly in videos making them inconsistent with language prompts. (2) While All-in-One is a unified framework for multi-modal VL tracking, it currently focuses mainly on language prompts. Actually, All-in-One has great potential to be extended to leverage more types of prompts, such as audio, point, mask, and scribble prompts~\cite{kirillov2023segment,zou2023segment}. We leave it for the future work.

\vspace{-0.05cm}
\myPara{Acknowledgement.} This work was supported by grants from the National Natural Science Foundation of China (No. 62101351), and the Key Research and Development Program of Chongqing (cstc2021jscx-gksbX0032).

\clearpage
\balance
\printbibliography
%\bibliographystyle{ACM-Reference-Format}
%\bibliography{egbib}

\end{document}